\documentclass{bmvc2k}
\usepackage{wrapfig}
\usepackage{amsmath,amssymb}
\usepackage{multirow}

\title{A mixed classification-regression framework for 3D pose estimation from 2D images}

\addauthor{Siddharth Mahendran}{siddharthm@jhu.edu}{1}
\addauthor{Haider Ali}{hali@jhu.edu}{2}
\addauthor{Ren\'{e} Vidal}{rvidal@cis.jhu.edu}{1}

\addinstitution{
Mathematical Institute for Data Science,\\
Johns Hopkins University,\\
Baltimore, MD, USA
}
\addinstitution{
Department of Computer Science,\\
Johns Hopkins University,\\
Baltimore, MD, USA
}

\runninghead{Mahendran \etal}{Mixed Classification Regression Framework}

\def\eg{\emph{e.g}\bmvaOneDot}
\def\ie{\emph{i.e}\bmvaOneDot}

\def\etal{\emph{et al}\bmvaOneDot}
\newcommand{\myparagraph}[1]{\smallskip\noindent\textbf{#1}.}

\begin{document}

\maketitle

\begin{abstract}
3D pose estimation from a single 2D image is an important and challenging task in computer vision with applications in autonomous driving, robot manipulation and augmented reality. Since 3D pose is a continuous quantity, a natural formulation for this task is to solve a \emph{pose regression} problem. However, since pose regression methods return a single estimate of the pose, they have difficulties handling multimodal pose distributions (\eg in the case of symmetric objects). An alternative formulation, which can capture multimodal pose distributions, is to discretize the pose space into bins and solve a \emph{pose classification} problem. However, pose classification methods can give large pose estimation errors depending on the coarseness of the discretization. In this paper, we propose a mixed classification-regression framework that uses a classification network to produce a discrete multimodal pose estimate and a regression network to produce a continuous refinement of the discrete estimate. The proposed framework can accommodate different architectures and loss functions, leading to multiple classification-regression models, so\-me of which achieve state-of-the-art performance on the challenging Pascal3D+ dataset.
\end{abstract}

\section{Introduction}
\label{sec:intro}

A fundamental problem in computer vision is to understand the underlying 3D geometry of the scene captured in a 2D image. This involves describing the scene in terms of the objects present in it (\ie object detection and classification) and predicting the relative rigid transformation between the camera and each object (\ie object pose estimation). The problem of estimating the 3D pose of an object from a single 2D image is an old problem in computer vision, which has seen renewed interest due to its applications in autonomous driving, robot manipulation and augmented reality, where it is very important to reason about objects in 3D. While general 3D pose estimation includes both the 3D rotation and translation between the object and the camera, here we restrict our attention to estimating only the 3D rotation.

Since 3D pose is a continuous quantity, a natural way to estimate it is to setup a pose regression problem: Given a dataset of 2D images of objects and their corresponding 3D pose annotations, the goal is to learn a regression function (\eg a deep network) that predicts the 3D pose of an object in an image. This requires choosing, \eg a network architecture, a representation for rotation matrices (Euler angles, axis-angles or quaternions) and a loss function (mean squared loss, Huber loss or geodesic loss). However, a disadvantage of regression-based approaches is that they are unimodal in nature (\ie they return a single pose estimate), hence they are unable to properly model multimodal distributions in the pose space 
which occur for object categories like boat and dining-table that exhibit strong~symmetries. 

An alternative approach that is able to capture multimodal distributions in the pose space is to setup a pose classification problem. Specifically, we discretize the pose space into bins and, instead of predicting a single pose output, we return a probability vector on the pose labels associated with these bins. This formulation is better at handling cases where two or more competing hypotheses have a high probability of being correct, for example due to symmetry. However, the drawback of this formulation is that we will always have a non-zero pose-estimation error even with perfect pose-classification accuracy due to the discretization process. Moreover, this error might be large if the binning is very coarse. 

In this paper, we propose a mixed classification-regression framework that combines the best of both worlds. The proposed framework consists of two main components. The first one is a classification network that predicts a pose label corresponding to a ``key pose'' obtained by discretizing the pose space. The second one is a regression network that predicts the deviation between the (continuous) object pose and the (discretized) key pose. The outputs of both components are then combined to predict the object pose. The proposed framework is fairly general and can accommodate multiple choices, for the pose classification and regression networks, for the way in which their outputs are combined, for the loss functions used to train them, etc. Experiments on the Pascal3D+ dataset show that our framework gives more accurate estimates of 3D pose than using either pure classification or regression.

\myparagraph{Related work} There are many non-deep learning methods for 3D pose estimation given 2D images. Due to space constraints, we restrict our review to only methods based on deep networks. The current literature on 3D pose estimation using deep networks can be divided in two groups: (i) methods that predict 2D keypoints from images and then recover the 3D pose from these keypoints,
and (ii) methods that directly predict 3D pose from an image. 

The first group of methods includes the works of \cite{Crivellaro:ICCV15,Grabner:CVPR18,Pavlakos:ICRA17,Wu:ECCV16,Rad:ICCV17}. The works of \cite{Pavlakos:ICRA17,Wu:ECCV16} train on 2D keypoints that correspond to semantic keypoints defined on 3D object models. Given a new image, they predict a probabilistic map of 2D keypoints and recover 3D pose by comparing with some pre-defined object models. In \cite{Grabner:CVPR18}, \cite{Rad:ICCV17} and \cite{Crivellaro:ICCV15}, instead of semantic keypoints, the 3D points correspond to the 8 corners of a 3D bounding box encapsulating the object. The network is trained by comparing the predicted 2D keypoint locations with the projections of the 3D keypoints on the image under ground-truth pose annotations. \cite{Grabner:CVPR18} uses a Huber loss on the projection error to be robust to inaccurate ground-truth annotations and is the current state-of-the-art on the Pascal3D+ dataset \cite{Xiang:WACV14} to the best of our knowledge.

The second group of methods includes the works of \cite{Tulsiani:CVPR15,Su:ICCV15,Elhoseiny:ICML16,Massa:BMVC16,Massa:arxiv14,Mahendran:ICCVW17,Wang:PCM16}. All these methods, except \cite{Mahendran:ICCVW17}, use the Euler angle representation of rotation matrices to estimate the azimuth, elevation and camera-tilt angles separately. \cite{Tulsiani:CVPR15}, \cite{Su:ICCV15} and \cite{Elhoseiny:ICML16} divide the angles into non-overlapping bins and solve a classification problem, while \cite{Wang:PCM16} tries to regress the angles directly with a mean squared loss. \cite{Massa:BMVC16} proposes multiple loss functions based on regression and classification and concludes that classification methods work better than regression ones. On the other hand, \cite{Mahendran:ICCVW17} uses the axis-angle and quaternion representations of 3D rotations and optimizes a geodesic loss on the space of rotation matrices. 

In this work we also use the axis-angle representation but within a mixed classification-regression framework (which we also call \emph{bin and delta model}) instead of the pure regression approach of \cite{Mahendran:ICCVW17}. The proposed framework can be seen as a generalization of \cite{Mousavian:CVPR17,Li:arxiv18,Guler:CVPR17,Guler:arxiv18}, which also use a bin and delta model to combine classification and regression networks. Specifically, \cite{Mousavian:CVPR17} is a variation of the \emph{geodesic bin and delta model} we propose in Eqn.~\eqref{eqn:m1+} with a $\cos-\sin$ representation of Euler angles, while \cite{Li:arxiv18} is a particular case of the \emph{simple bin and delta model} we propose in Eqn.~\eqref{eqn:m0+} with a quaternion representation of 3D pose. On the other hand, the quantized regression model of \cite{Guler:CVPR17,Guler:arxiv18} uses the bin and delta model to generate dense correspondences between a 3D model and an image for face landmark and human pose estimation. \cite{Guler:CVPR17} learns a modification of our \emph{simple bin and delta model}  in Eqn.~\eqref{eqn:m0} with a separate delta network for every facial region while \cite{Guler:arxiv18} is a particular case of the \emph{probabilistic bin and delta model} we propose in Eqn.~\eqref{eqn:m3}. \cite{Guler:arxiv18} also makes a connection between a bin and delta model and a mixture of regression experts proposed in \cite{Jordan:NC94}, where the classification output probability vector acts as a gating function on regression experts.

Broadly speaking, there has been recent interest in trying to design networks and representations that combine classification and regression to model 3D pose but the authors of these different works have treated this as a one-off representation problem. In contrast, we propose a general framework that encapsulates prior models as particular cases.

\myparagraph{Paper outline} The remainder of the paper is organized as follows. In \S\ref{sec:problem} we define the 3D pose estimation problem along with our network architecture, 3D pose representation and some common-sense baselines that use only regression or classification. In \S\ref{sec:framework} we describe our mixed classification-regression framework and discuss a variety of different models and loss functions that arise from the framework. Finally, in \S\ref{sec:results} we demonstrate the effectiveness of these models on the challenging Pascal3D+ dataset for the task of 3D pose estimation. 

\section{3D Pose Estimation}
\label{sec:problem}
\begin{figure}
\centering
\subfigure[Overview of the problem]{\includegraphics[width=0.5\linewidth]{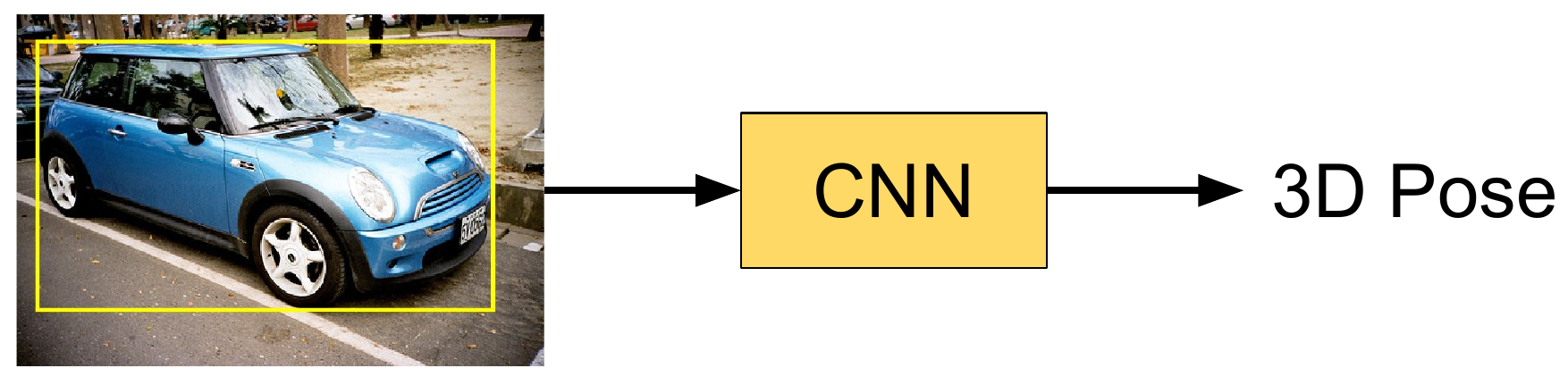} \label{fig:problem_statement}}
~
\subfigure[Overview of the network architecture]{\includegraphics[width=0.4\linewidth]{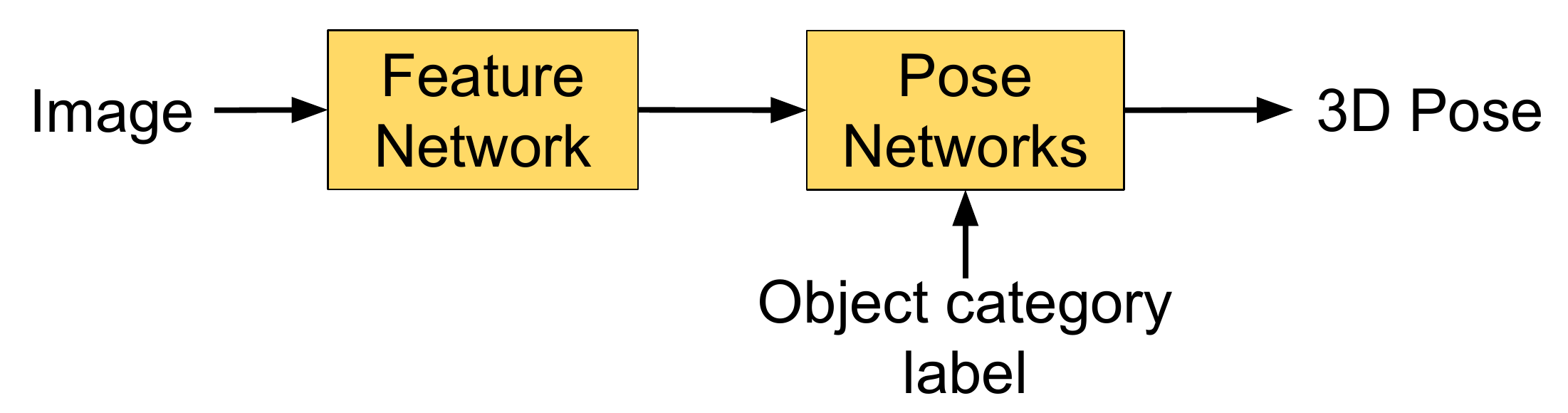} \label{fig:network_architecture}}
\caption{A high level overview of our problem statement and proposed network architecture}
\vspace{-3mm}
\end{figure}

Given an image and a bounding box around an object in the image with known object category label, in this paper we consider the problem of estimating its 3D pose. We assume the bounding box around the object is given by an oracle, but the output of an object detection system like \cite{Ren:FasterRCNN} can also be used instead. An overview of our problem is shown in Fig.~\ref{fig:problem_statement}.

\myparagraph{Network architecture} We use a standard network architecture shown in Fig.~\ref{fig:network_architecture} where we have a feature network shared across all object categories and a pose network per object category. The final pose output is selected as the output of the pose network corresponding to the input object category label. We use the ResNet-50 network \cite{He:CVPR16,He:arxiv16} minus the last classification layer as our feature network. Each pose network is essentially a multi-layer perceptron, with fully connected layers containing ReLU nonlinearities and Batch-Normalization layers in between, whose output depends on 
whether it is a classification, regression or mixed pose network. We describe the proposed bin and delta pose network in more detail in \S\ref{sec:framework}.

\myparagraph{Representations of 3D pose} Let $SO(3) = \{ R \in \mathbb{R}^{3 \times 3} : R^T R = I, ~ \det(R) = +1 \}$ denote the Special Orthogonal group of rotation matrices of dimension 3. We use the axis-angle representation $R = \exp ( \theta [v]_{\times})$ of a rotation matrix $R\in SO(3)$ with the corresponding axis-angle vector defined as $y = \theta v$, where $\theta$ is the angle of rotation, $v = [v_1, v_2, v_3]^T$ ($\|v\|_2 = 1$) is the axis of rotation, $[v]_\times = [[0, -v_3, v_2], [v_3, 0, -v_1], [-v_2, v_1, 0]] $, and $\exp$ is the matrix exponential. Assuming $\theta \in [0, \pi)$ and defining $R = I_3$ iff $y=0$ sets up a bijective mapping between axis-angle vector $y$ and rotation matrix $R$. 
With the above notation, we can write the input-output map of the network architecture in Fig.~\ref{fig:network_architecture} as $y = \Phi(x; W)$, where $x$ denotes the input image, $W$ denotes the network weights, and $y$ denotes the axis angle representation. Technically, the output $y$ is also a function of the object category label $c$, \ie $y=\Phi(x, c; W)$. For the sake of simplicity, we will omit this detail in further analysis.

\myparagraph{Regression baseline} A natural baseline is a regression formulation with a squared Euclidean loss between the ground-truth 3D pose $y_n^*$ and the predicted pose $y_n = \Phi_R(x_n; W)$ for image $x_n$. This involves solving the following optimization problem $\mathcal{R}_E$ during training:
\begin{equation}
\mathcal{R}_E: \hspace{1cm} \min_{W} \frac{1}{N} \sum_n \| y_n^* - \Phi_R(x_n; W) \|_2^2.
\label{eqn:r0}
\end{equation}
However, as recommended by \cite{Mahendran:ICCVW17}, it makes more sense to minimize a geodesic loss on the space of rotation matrices instead of the Euclidean loss as it better captures the geometry of the problem. This involves solving the following optimization problem $\mathcal{R}_G$ during training:
\begin{equation}
\mathcal{R}_G: \hspace{1cm} \min_{W} \frac{1}{N} \sum_n \mathcal{L}_p(y_n^*, \Phi_R(x_n; W)),
\label{eqn:r1}
\end{equation}
where $\mathcal{L}_p(y_1, y_2) \equiv \mathcal{L}(R_1, R_2) = \frac{1}{\sqrt{2}}\|\log(R_1^T R_2)\|_F$ is the geodesic distance between two axis-angle vectors $y_1$ and $y_2$ with corresponding rotation matrices $R_1$ and $R_2$ respectively. 

\myparagraph{Classification baseline} An alternative to the regression baselines $\mathcal{R}_E$ and $\mathcal{R}_G$ is a classification baseline $\mathcal{C}$. Here, we first run K-Means clustering on the ground-truth pose annotations $\{y_n^*\}_{n=1}^N$ to obtain two things: (i) a pose label $l_n^*$ associated with every image $x_n$ and (ii) a K-Means dictionary $\{z_k\}_{k=1}^K$ which also acts as the set of key poses in the discretization process. We then train a network to predict the pose labels by minimizing the cross-entropy loss $\mathcal{L}_c$ between the ground-truth pose labels $l_n^*$ and the predicted pose labels $\Phi_C(x_n; W)$:
\begin{equation}
\mathcal{C}: \hspace{1cm} \min_W \frac{1}{N} \sum_n \mathcal{L}_c(l_n^*, \Phi_C(x_n; W)).
\label{eqn:c0}
\end{equation}

\section{Mixed Classification-Regression Framework}
\label{sec:framework}

This section presents the proposed classification-regression framework. \S\ref{sec:overview} presents  an~over-view of the Bin \& Delta model and its network architecture, while  \S\ref{sec:m0}-\S\ref{sec:m3} present various loss functions that arise from different modeling choices within our general framework. 

\subsection{Overview of the Bin \& Delta model}
\label{sec:overview}

\begin{wrapfigure}{r}{0.5\linewidth}
	\includegraphics[width=\linewidth]{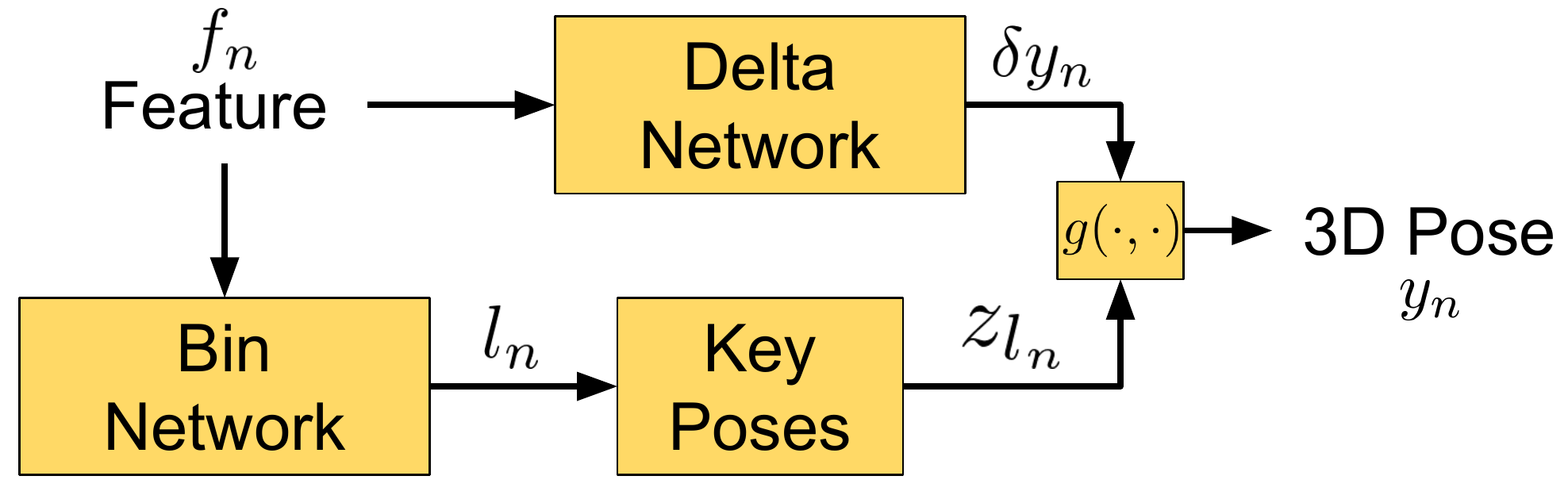}
	\caption{Overview of the Bin\&Delta model}
	\label{fig:bin_and_delta}
\end{wrapfigure}

Instead of a single multi-layer perceptron as the pose network like we have in our regression and classification baselines, the Bin \& Delta model has two components: a bin network and a delta network as shown in Fig.~\ref{fig:bin_and_delta}. They both take as input the output of the feature network, $f_n = \Phi_F(x_n; W_F)$, where $\Phi_F$ is the feature network parameterized by weights $W_F$ and $x_n$ is the input image. Given feature input $f_n$, the bin network predicts a pose label, $l_n = \Phi_B(f_n; W_B)$, where $\Phi_B$ is the bin network parameterized by weights $W_B$. 
The pose label $l_n$ references a key pose $z_{l_n}$ in the discretization process. The delta network predicts a pose residual, $\delta y_n = \Phi_D(f_n; W_D)$, where $\Phi_D$ is the delta network parameterized by weights $W_D$. The discrete pose $z_{l_n}$ and the pose residual $\delta y_n$ are combined to predict the continuous pose $y_n$ as 
\begin{equation}
y_n = g(z_{l_n}, \delta y_n),
\label{eqn:b&d}
\end{equation}
where different choices for $g$ lead to different ways of combining the outputs of the bin and delta models. We denote all the parameters of the bin and delta model as $W = [W_F, W_B, W_D]$. 
 
The above framework for combining classification and regression is very general and there are many design choices that lead to different models and loss functions. For example:
\begin{enumerate}
\item \textit{How to combine the classification and regression outputs?} Choosing the function $g$ to be the addition operation, \ie $y_n = z_{l_n} + \delta y_n$, leads to our models in \S\ref{sec:m0}, \S\ref{sec:m1} and \S\ref{sec:m3}. Alternatively, taking the log of the product of the rotations associated to the outputs of the bin and delta models, \ie $y_n = \log ( \exp(z_{l_n}) \exp(\delta y_n))$, leads to our model in \S\ref{sec:m2}. 

\item \textit{Where to apply the regression loss?} We can choose to provide supervision during training at either the final pose output or at the intermediate output of the delta network. The former leads to our model in \S\ref{sec:m1} and the later leads to our model in \S\ref{sec:m0}. 

\item \textit{Hard or soft assignment in the pose-binning step?} Instead of assigning a single pose label for every image (a hard assignment), we can assign a probability vector over pose-bins (a soft assignment). This leads to our model in \S\ref{sec:m3}. 

\item \textit{Single delta network for all pose-bins or one per pose-bin?} This is a decision choice we can make for all the models in \S\ref{sec:m0}-\S\ref{sec:m3} and we discuss it in \S\ref{sec:multiple}.
\end{enumerate}
Also, note that even though we present everything in the context of axis-angle representation of 3D pose, all our proposed models can be generalized to any choice of pose representation. 

\subsection{Simple/Naive Bin \& Delta}
\label{sec:m0}
Given training data $\{x_n, y_n^*\}_{n=1}^N$, of images $x_n$ and corresponding ground-truth pose-targets $y_n^*$, we run the K-Means discretization process outlined in the classification baseline to associate a pose label $l_n^*$ with every image. Given this label and the key poses $\{z_k\}_{k=1}^K$, we can obtain a ground-truth delta $\delta y_n^* = y_n^* - z_{l_n^*}$ for every image. 
Now, the bin and delta networks can be trained on modified training data $\{x_n, l_n^*, \delta y_n^* \}_{n=1}^N$ with a cross-entropy loss $\mathcal{L}_c$ for the bin network and a Euclidean loss $\| \cdot \|_2^2$ for the delta network. More specifically, the parameters $W$ are learned by solving the following optimization problem: 
\begin{equation}
\mathcal{M}_S: \hspace{1cm} \min_W \frac{1}{N} \sum_n \left [ \mathcal{L}_c(l_n^*, l_n) + \alpha \|\delta y_n^* - \delta y_n \|_2^2 \right ],
\label{eqn:m0}
\end{equation}
where $\alpha \geq 0$ is a relative weighting parameter for balancing the two losses.

\subsection{Geodesic Bin \& Delta}
\label{sec:m1}
The Simple Bin \& Delta model penalizes incorrect predictions in the individual bin and delta networks. It is not cognizant of the fact that what we care about is the final predicted pose. To address this issue, we propose a new Bin \& Delta model that regresses the final output pose instead of the intermediate output of the delta network. We call this a Geodesic Bin \& Delta model because we apply a geodesic regression loss $\mathcal{L}_p$ between the ground-truth pose and predicted pose by solving the following optimization problem:
\begin{equation}
\mathcal{M}_G: \hspace{1cm} \min_W \frac{1}{N} \sum_n \left [ \mathcal{L}_c(l_n^*, l_n) + \alpha  \mathcal{L}_p(y_n^*, z_{l_n} + \delta y_n) \right ].
\label{eqn:m1}
\end{equation}
Notice that that this model has strong connections to the regression baseline $\mathcal{R}_G$, except that we now model multimodal 3D pose-distributions and have an additional classification loss. As noted in \cite{Mahendran:ICCVW17}, the geodesic loss is non-convex with many local minima and a good initialization is required. We initialize the networks by training on problem $\mathcal{M}_S$ for 1 epoch. 

\subsection{Riemannian Bin \& Delta}
\label{sec:m2}

\begin{wrapfigure}{r}{0.3\linewidth}
\centering
\vspace{-1cm}
\includegraphics[width=\linewidth]{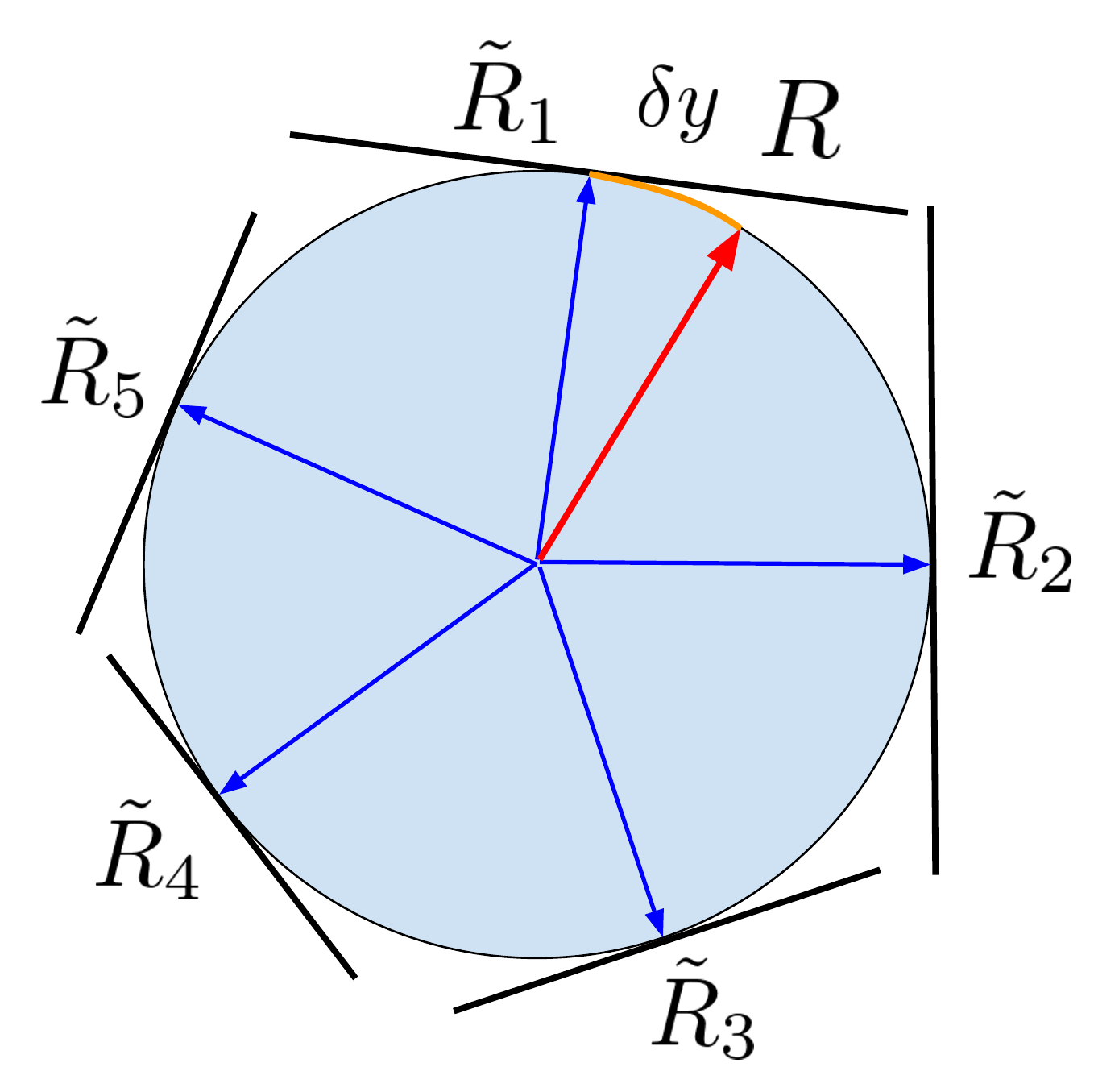}
\vspace{-8mm}
\caption{Toy example of the Riemannian bin \& delta model for a circle}
\label{fig:rotation_bin_delta}
\vspace{-2mm}
\end{wrapfigure}
In the two models discussed so far, we have assumed that $\delta y_n = y_n - z_{l_n}$, which does not truly capture the geometry of the rotation group $SO(3)$. Technically, the delta model $\delta y_n$ should capture the deviations between the continuous pose $R_n$ and the discrete pose $\tilde{R}_k = \exp(z_k)$ associated with key pose $z_k$. Specifically, $\delta y_n$ should be a tangent vector at $\tilde{R}_k$ in the direction of $R_n$ whose norm is equal to the geodesic distance between $\tilde{R}_k$  and $R_n$. Mathematically, this is expressed using exponential and logarithm maps as:
\begin{equation}
\delta y_n = \log_{\tilde{R}_{l_n}}(R_n) = \log \left ( \tilde{R}_{l_n}^T R_n \right ) ~\text{or}~~  R_n = \tilde{R}_{l_n}\exp(\delta y_n).
\label{eqn:rot_delta}
\end{equation}
The geodesic loss between the ground-truth and predicted rotations $\mathcal{L}(R_n^*, \tilde{R}_{l_n} \exp(\delta y_n)) = \mathcal{L}(\tilde{R}_{l_n}^T R_n^*, \exp(\delta y_n))$ can be approximated by the Euclidean distance on the tangent space at the identity as $\| \log (\tilde{R}_{l_n}^T R_n^*) - \log (\exp (\delta y_n)) \|_2$. This new regression loss gives us the Riemannian Bin \& Delta model, which is based on solving the following optimization problem 
\begin{align}
\mathcal{M}_R: \hspace{1cm} & \min_W \frac{1}{N} \sum_n \left [ \mathcal{L}_c(l_n^*, l_n) + \alpha  \|\log(\tilde{R}_{l_n}^T R_n^*) - \delta y_n \|_2^2 \right ] ,
\label{eqn:m2} 
\end{align}
where the term $\log(\tilde{R}_{l_n}^T R_n^*)$ can be precomputed for efficiency of training.
Fig.~\ref{fig:rotation_bin_delta} shows a toy example of our proposed model in the context of a circle. We show a circle with 5 tangent planes corresponding to key poses $\tilde{R}_i, i=1,...,5$. The rotation $R$ (shown in red) is now a combination of the key pose $\tilde{R}_1$, with pose label $l=1$, and the delta $\delta y$ (shown in orange).

\subsection{Probabilistic Bin \& Delta}
\label{sec:m3}

In all the models discussed so far we have used a deterministic (hard) assignment obtained from K-Means in which we assign a single key-pose to an image. A more flexible and possibly more informative model would be to do a probabilistic (soft) assignment to all key-poses. Specifically, post K-Means we can generate a probabilistic assignment as:
\vspace{-1mm}
\begin{equation}
p_{nk}^* = \frac{\exp(- \gamma \|y_n^* - z_k\|_2^2)}{\sum_k \exp(- \gamma \|y_n^* - z_k\|_2^2)}.
\label{eqn:soft}
\vspace{-1mm}
\end{equation}
Now, the classification loss can be modified to be a Kullback-Leibler (KL) divergence between ground-truth and predicted probabilities. 
The regression loss is also updated to fully utilize this probabilistic output as shown in the following optimization problem: 
\begin{equation}
\mathcal{M}_P: \hspace{1cm} \min_W \frac{1}{N} \sum_n \left [\mathcal{L}_{KD}(p_n^*, p_n) + \alpha \sum_k p_{nk} \mathcal{L}_p(y_n^*, z_k + \delta y_n) \right ] .
\label{eqn:m3}
\end{equation}
The predicted pose is now $y_n = z_{l_n} + \delta y_n$, where $l_n = \operatorname{argmax}_k p_{nk}$. Another variation that is of relevance here is that instead of using K-Means followed by the probabilistic assignment of Eqn.~\eqref{eqn:soft}, one could learn a Gaussian Mixture Model (GMM) in the pose-space to do the soft assignment in a more natural way. We did not explore this variation but mention it to demonstrate that many more models can be described as particular cases of our framework. 

\vspace{-2mm}
\subsection{One delta network per pose-bin}
\label{sec:multiple}

\begin{wrapfigure}{r}{0.45\linewidth}
\vspace{-0.5cm}
\includegraphics[width=\linewidth]{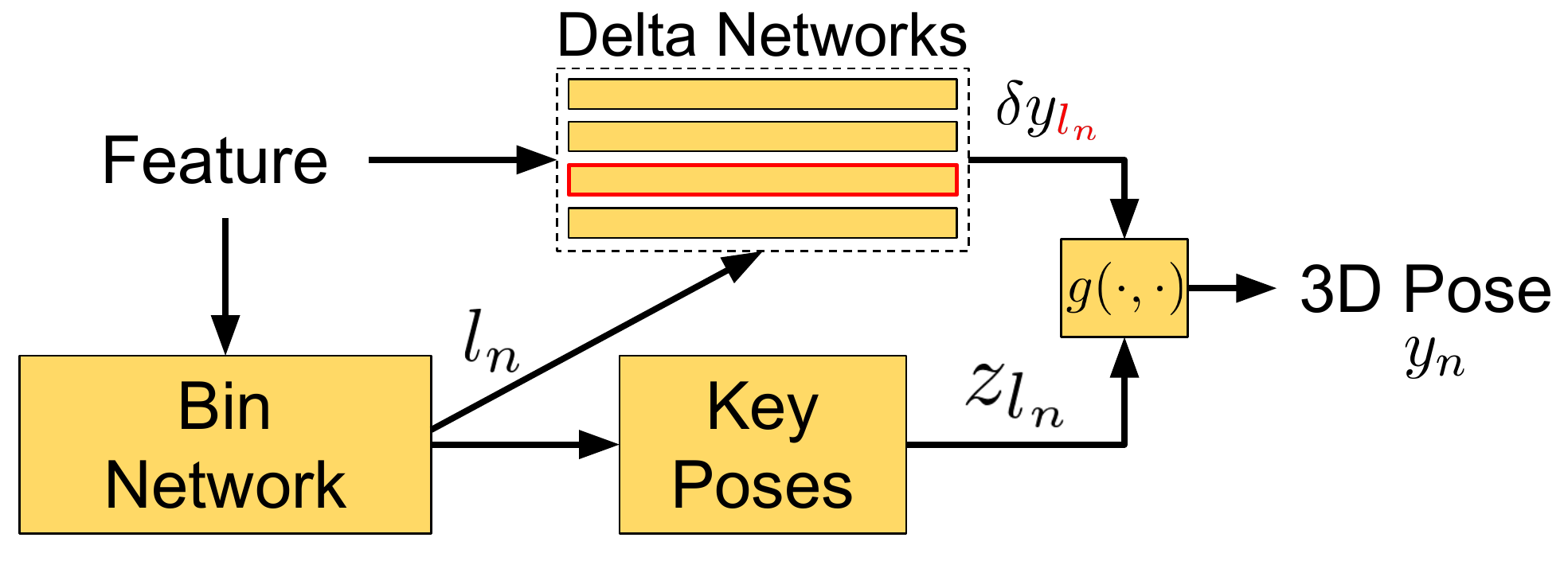}
\caption{One delta network per pose-bin}
\label{fig:multiple_delta}
\vspace{-0.3cm}
\end{wrapfigure}
An implicit assumption in all the models discussed so far is that there is a single delta model for all pose-bins. An alternative modeling choice, as shown in Fig.~\ref{fig:multiple_delta}, is to have a delta model for every single pose-bin. This modeling decision is equivalent to deciding whether to have a common covariance matrix across all clusters or have a different covariance matrix for every cluster in a GMM. If we choose to now have one delta network per pose-bin, we can update all the previous optimization problems as follows (the change is highlighted in red):
\vspace{-2mm}
\begin{align}
\mathcal{M}_S+: \hspace{1cm} & \min_W \frac{1}{N} \sum_n \left [ \mathcal{L}_c(l_n^*, l_n) + \alpha \|\delta y_n^* - \delta y_{\textcolor{red}{l_n}} \|_2^2 \right ],
\label{eqn:m0+} \\
\mathcal{M}_G+: \hspace{1cm} & \min_W \frac{1}{N} \sum_n \left [ \mathcal{L}_c(l_n^*, l_n) + \alpha  \mathcal{L}_p(y_n^*, z_{l_n} + \delta y_{\textcolor{red}{l_n}}) \right ],
\label{eqn:m1+} \\
\mathcal{M}_R+: \hspace{1cm} & \min_W \frac{1}{N} \sum_n \left [ \mathcal{L}_c(l_n^*, l_n) + \alpha \|\log(\tilde{R}_{l_n}^T R_n^*) - \delta y_{\textcolor{red}{l_n}} \|_2^2 \right ],
\label{eqn:m2+} \\
\mathcal{M}_P+: \hspace{1cm} & \min_W \frac{1}{N} \sum_n \left [\mathcal{L}_{KD}(p_n^*, p_n) + \alpha \sum_k p_{nk} \mathcal{L}_p(y_n^*, z_k + \delta y_{\textcolor{red}{nk}}) \right ] .
\label{eqn:m3+}
\vspace{-5mm}
\end{align}

\vspace{-8mm}
\section{Results}
\label{sec:results}

First, we describe the Pascal3D+ dataset \cite{Xiang:WACV14}, which is the benchmark dataset used for evaluating 3D pose estimation methods. Then, we demonstrate the effectiveness of our framework with state-of-the-art performance on this challenging task. Finally, we present an ablation study on two hyper-parameters: size of the K-Means dictionary $K$ and relative weight $\alpha$.

\myparagraph{Dataset} The Pascal3D+ consists of images of twelve object categories: aeroplane (aero), bicycle (bike), boat, bottle, bus, car, chair, diningtable (dtable), motorbike (mbike), sofa, train and tvmonitor (tv). These images were curated from the Pascal VOC \cite{PASCAL} and ImageNet \cite{ImageNet} datasets, and annotated with 3D pose in terms of the Euler angles $(az, el, ct)$. We use the ImageNet-trainval and Pascal-train images as our training data and the Pascal-val images as our testing data. Following the protocol of \cite{Tulsiani:CVPR15,Su:ICCV15} and others, we use ground-truth bounding boxes of un-occluded and un-truncated objects. We use the 3D pose-jittering data augmentation strategy of \cite{Mahendran:ICCVW17} and the rendered images of \cite{Su:ICCV15} to augment our training data. 

\myparagraph{Evaluation metrics} We evaluate our models on the Pascal3D+ dataset under two standard metrics: (i) median angle error across all test images $MedErr$ and (ii) percentage of images that have angle error less than $\frac{\pi}{6}$, $Acc_{\frac{\pi}{6}}$. Here the angle error is the angle between predicted and ground-truth rotation matrices given by $\frac{1}{\sqrt{2}} \| \log \left ( R_1^T R_2 \right ) \|_F$. For all the results, we run each experiment three times and report the mean and standard deviation across three trials. 

\begin{figure}
\centering
\subfigure[$MedErr$ metric (lower is better)]{\includegraphics[width=0.45\linewidth]{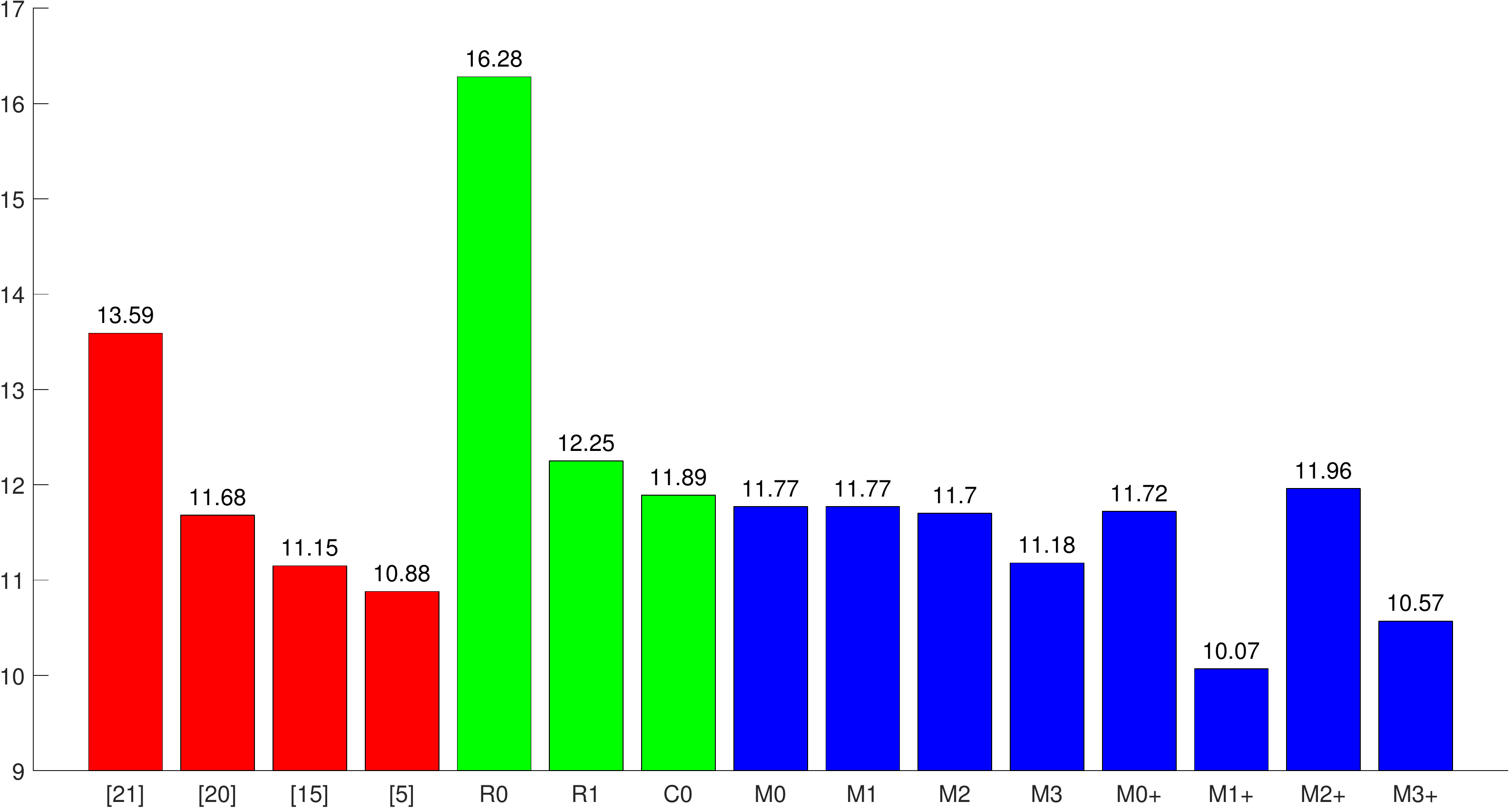} \label{fig:results_mederr}}
~
\subfigure[$Acc_{\frac{\pi}{6}}$ metric (higher is better)]{\includegraphics[width=0.45\linewidth]{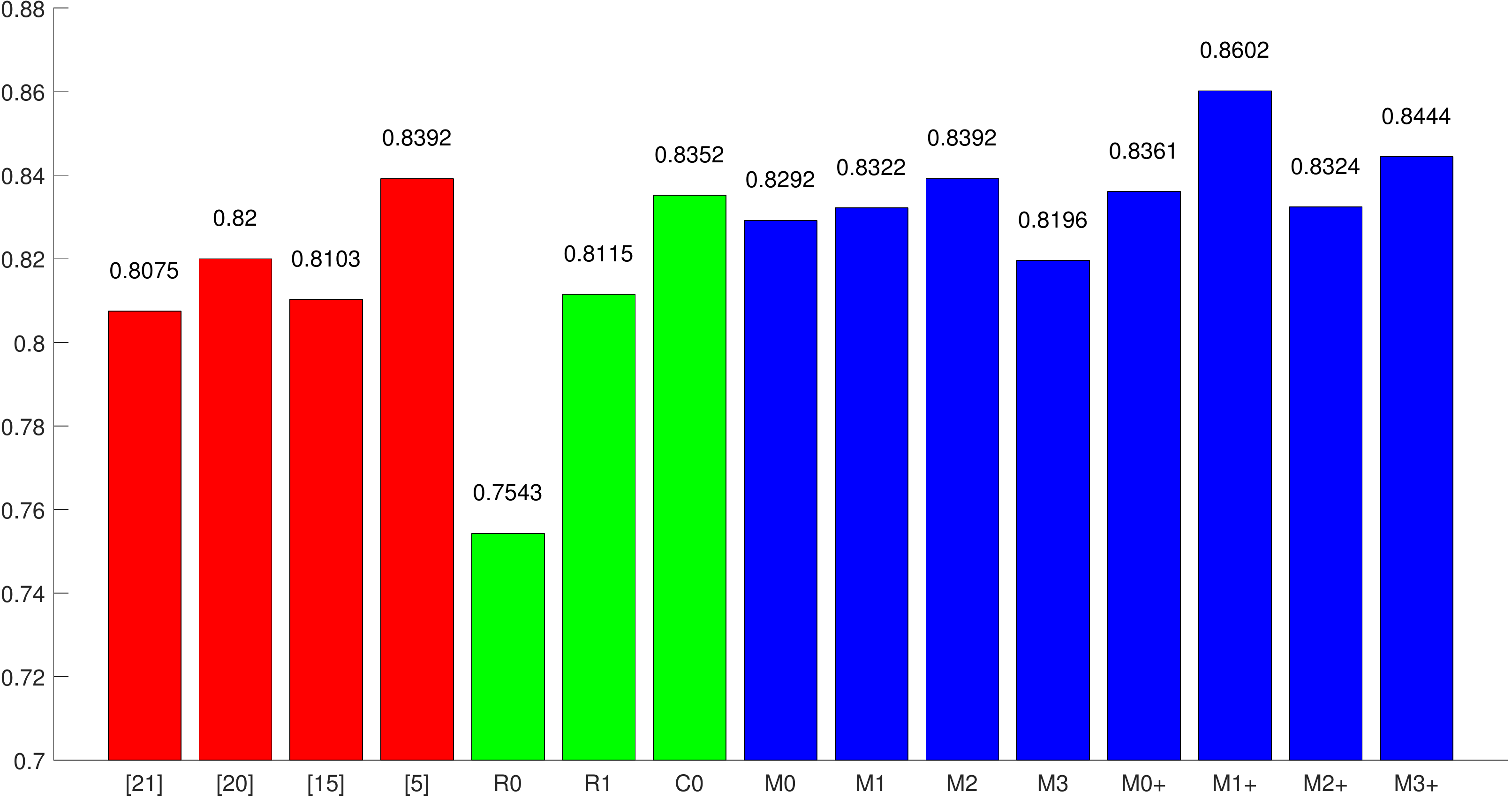} \label{fig:results_acc}}
\caption{Performance on the Pascal3D+ dataset under two metrics (averaged over all object categories). Red bars are current state-of-the-art methods, green bars are our baselines and blue bars are our proposed models (best seen in color)}
\label{fig:results}
\end{figure}

\myparagraph{3D pose estimation using Bin \& Delta models}
As can be seen in Tables~\ref{table:MedErr}, \ref{table:Acc30} and Fig.~\ref{fig:results}, we achieve state-of-the-art performance with our Geodesic and Probabilistic bin \& delta models with one delta network per pose-bin, namely models $\mathcal{M}_G+$ and $\mathcal{M}_P+$, respectively. We improve upon the existing state-of-the-art \cite{Grabner:CVPR18}, from $10.88^\circ$ to $10.10^\circ$ in the $MedErr$ metric and from 0.8392 to 0.8588 in the $Acc_{\frac{\pi}{6}}$ metric under model $\mathcal{M}_G+$. 
Other important observations include: (1) Models $\mathcal{M}_G+$ and $\mathcal{M}_P+$ that have one delta network per pose-bin perform better than the models $\mathcal{M}_G$ and $\mathcal{M}_P$ with a single delta network for all pose-bins, as they have more freedom in modeling the deviations from key pose even though the discretization of the pose space is coarser. (2) Models that use a geodesic loss on the final pose output ($\mathcal{M}_G$, $\mathcal{M}_P$, $\mathcal{M}_G+$ and $\mathcal{M}_P+$) perform better than the pure classification \& regression baselines $\mathcal{C}, \mathcal{R}_E, \mathcal{R}_G$ as well as models that put losses on the individual components ($\mathcal{M}_S$, $\mathcal{M}_R$, $\mathcal{M}_S+$ and $\mathcal{M}_P+$). (3) Pure regression with a geodesic loss $\mathcal{R}_G$ worked better than pure classification $\mathcal{C}$ under the $MedErr$ metric but worse under the $Acc_\frac{\pi}{6}$ metric. (4) Geodesic regression $\mathcal{R}_G$ is better than Euclidean regression $\mathcal{R}_E$.

\myparagraph{Ablation analysis} There are two main hyper-parameters in our model, (i) the size of the K-Means dictionary $K$ and (ii) the relative weighting parameter $\alpha$. In Tables~\ref{table:kmeans_m1} and \ref{table:kmeans_m1+}, we vary the size of the K-Means dictionary for models $\mathcal{M}_G$ $(K= \{24, 50, 100, 200\})$ and $\mathcal{M}_G+$ $(K = \{4, 8, 16, 24 \})$ respectively. We see that usually, a larger K-Means dictionary is better. We also note that even with a very coarsely discretized pose space ($K=4$), $\mathcal{M}_G+$ is better than the very highly discretized pose space ($K=200$) of $\mathcal{M}_G$. In Tables~\ref{table:alpha_m1} and \ref{table:alpha_m1+}, we tried three different values of $\alpha=\{0.1, 1, 10\}$ for models $\mathcal{M}_G$ and $\mathcal{M}_G+$ respectively and observed that performance improves with higher $\alpha$. This is because higher $\alpha$ gives more importance to the geodesic loss on the final predicted pose which is the quantity we care about.

\section{Conclusion}
\label{sec:conclusion}

3D pose estimation is an important but challenging task and current deep learning solutions solve this problem using pure regression or classification approaches. We designed a framework that combines regression and classification loss functions to predict fine-pose estimates while modeling multi-modal pose distributions. This was implemented using a flexible bin and delta network architecture where different modeling choices led to different models and loss functions. We analyzed these models on the Pascal3D+ dataset and demonstrated state-of-the-art performance using two of them. We also tested various modeling choices and provided some guidelines for future work using these models. 

\myparagraph{Acknowledgement} This research was supported by NSF grant 1527340.

\begin{table}
\centering
\small
\setlength{\tabcolsep}{1.2mm}
\begin{tabular}{|c|cccccccccccc|c|}
\hline
Method & aero & bike & boat & bottle & bus & car & chair & dtable & mbike & sofa & train & tv & Mean \\
\hline
\cite{Tulsiani:CVPR15} & 13.8 & 17.7 & 21.3 & 12.9 & 5.8 & 9.1 & 14.8 & 15.2 & 14.7 & 13.7 & 8.7 & 15.4 & 13.59 \\
\cite{Su:ICCV15} & 15.4 & \textcolor{red}{14.8} & 25.6 & 9.3 & 3.6 & 6.0 & \textcolor{red}{9.7} & \textbf{10.8} & 16.7 & \textbf{9.5} & 6.1 & 12.6 & 11.68 \\
\cite{Mousavian:CVPR17} & 13.6 & \textbf{12.5} & 22.8 & 8.3 & 3.1 & 5.8 & 11.9 & 12.5 & \textcolor{red}{12.3} & 12.8 & 6.3 & 11.9 & 11.15 \\
\cite{Grabner:CVPR18} & \textcolor{red}{10.0} & 15.6 & \textbf{19.1} & 8.6 & 3.3 & \textcolor{red}{5.1} & 13.7 & 11.8 & \textbf{12.2} & 13.5 & 6.7 & \textbf{11.0} & 10.88 \\
\hline
$\mathcal{R}_E$ & 14.5 & 17.7 & 39.3 & 7.4 & 4.0 & 7.8 & 15.2 & 26.6 & 17.5 & 10.5 & 11.5 & 14.1 & 15.50 \\ 
$\mathcal{R}_G$ & 11.8 & 15.9 & 27.2 & 7.2 & 2.9 & 5.2 & 11.6 & 15.0 & 14.3 & 10.8 & \textcolor{red}{5.4} & 12.4 & 11.63 \\ 
$\mathcal{C}$ & 11.7 & 15.3 & 21.5 & 9.3 & 4.1 & 7.4 & 11.2 & 17.8 & 17.0 & 11.0 & 7.0 & 13.1 & 12.20 \\ 
\hline
$\mathcal{M}_S$ & 11.0 & 15.5 & 21.0 & 8.8 & 3.8 & 7.0 & 10.8 & 21.0 & 16.6 & 10.7 & 6.5 & 13.1 & 12.14 \\ 
$\mathcal{M}_G$ & 10.6 & 16.4 & 21.6 & 8.1 & 3.2 & 6.0 & 9.9 & 14.6 & 16.0 & 11.1 & 6.3 & 13.4 & 11.44 \\  
$\mathcal{M}_R$ & 12.8 & 15.2 & 23.4 & 9.0 & 4.0 & 7.4 & 11.1 & 16.8 & 16.1 & 10.7 & 6.6 & 12.3 & 12.11 \\ 
$\mathcal{M}_P$ & 11.4 & 16.3 & 25.6 & \textcolor{red}{7.0} & \textbf{2.6} & 5.1 & 11.3 & 16.0 & 13.6 & 10.2 & 5.5 & 12.0 & 11.38 \\ 
\hline
$\mathcal{M}_S+$ & 12.2 & 15.7 & 24.4 & 9.9 & 3.6 & 6.5 & 12.0 & 14.8 & 14.4 & 11.9 & 6.4 & \textcolor{red}{11.6} & 11.95 \\ 
$\mathcal{M}_G+$ & \textbf{8.5} & \textcolor{red}{14.8} & \textcolor{red}{20.5} & \textcolor{red}{7.0} & 3.1 & \textcolor{red}{5.1} & \textbf{9.3} & \textcolor{red}{11.3} & 14.2 & 10.2 & 5.6 & 11.7 & \textbf{10.10} \\  
$\mathcal{M}_R+$ & 12.3 & 16.7 & 24.7 & 7.5 & 3.6 & 6.5 & 11.5 & 15.5 & 15.1 & 11.1 & 7.3 & 12.1 & 11.99 \\ 
$\mathcal{M}_P+$ & 10.6 & 15.0 & 23.9 & \textbf{6.7} & \textcolor{red}{2.7} & \textbf{4.7} & 9.8 & 12.6 & 13.9 & \textcolor{red}{9.7} & \textbf{5.3} & 11.7 & \textcolor{red}{10.54} \\ 
\hline
\end{tabular}
\caption{Performance of our models under the $MedErr$ metric (lower is better). Best results are highlighted in bold and the second best are shown in red (best seen in color).}
\label{table:MedErr}
\end{table}

\begin{table}
\centering
\footnotesize
\setlength{\tabcolsep}{1.4mm}
\begin{tabular}{|c|cccccccccccc|c|}
\hline
Method & aero & bike & boat & bottle & bus & car & chair & dtable & mbike & sofa & train & tv & Mean \\
\hline
\cite{Tulsiani:CVPR15} & 0.81 & 0.77 & 0.59 & 0.93 & 0.98 & 0.89 & 0.80 & 0.62 & 0.88 & 0.82 & 0.80 & 0.80 & 0.8075 \\
\cite{Su:ICCV15} & 0.74 & 0.83 & 0.52 & 0.91 & 0.91 & 0.88 & 0.86 & 0.73 & 0.78 & 0.90 & 0.86 & 0.92 & 0.8200 \\
\cite{Mousavian:CVPR17} & 0.78 & 0.83 & 0.57 & 0.93 & 0.94 & 0.90 & 0.80 & 0.68 & 0.86 & 0.82 & 0.82 & 0.85 & 0.8103 \\
\cite{Grabner:CVPR18} & 0.83 & 0.82 & 0.64 & 0.95 & 0.97 & 0.94 & 0.80 & 0.71 & 0.88 & 0.87 & 0.80 & 0.86 & 0.8392 \\
\hline
$\mathcal{R}_E$ & 0.77 & 0.75 & 0.41 & 0.96 & 0.91 & 0.83 & 0.72 & 0.56 & 0.75 & 0.90 & 0.75 & 0.87 & 0.7656 \\ 
$\mathcal{R}_G$ & 0.80 & 0.78 & 0.54 & 0.97 & 0.95 & 0.93 & 0.83 & 0.59 & 0.82 & 0.91 & 0.81 & 0.86 & 0.8166 \\ 
$\mathcal{C}$ & 0.84 & 0.77 & 0.60 & 0.95 & 0.97 & 0.95 & 0.90 & 0.63 & 0.78 & 0.94 & 0.81 & 0.87 & 0.8350 \\ 
\hline
$\mathcal{M}_S$ & 0.83 & 0.78 & 0.61 & 0.96 & 0.96 & 0.94 & 0.90 & 0.56 & 0.79 & 0.95 & 0.82 & 0.87 & 0.8303 \\ 
$\mathcal{M}_G$ & 0.84 & 0.76 & 0.62 & 0.96 & 0.98 & 0.94 & 0.92 & 0.65 & 0.80 & 0.96 & 0.82 & 0.87 & 0.8439 \\ 
$\mathcal{M}_R$ & 0.83 & 0.77 & 0.58 & 0.96 & 0.96 & 0.94 & 0.91 & 0.71 & 0.81 & 0.93 & 0.81 & 0.87 & 0.8410 \\ 
$\mathcal{M}_P$ & 0.80 & 0.77 & 0.56 & 0.97 & 0.97 & 0.93 & 0.82 & 0.57 & 0.81 & 0.92 & 0.82 & 0.88 & 0.8185 \\ 
\hline
$\mathcal{M}_S+$ & 0.82 & 0.80 & 0.59 & 0.94 & 0.97 & 0.94 & 0.91 & 0.63 & 0.81 & 0.97 & 0.83 & 0.87 & 0.8387 \\ 
$\mathcal{M}_G+$ & 0.87 & 0.81 & 0.64 & 0.96 & 0.97 & 0.95 & 0.92 & 0.67 & 0.85 & 0.97 & 0.82 & 0.88 & 0.8588 \\ 
$\mathcal{M}_R+$ & 0.81 & 0.77 & 0.56 & 0.96 & 0.97 & 0.92 & 0.86 & 0.73 & 0.79 & 0.93 & 0.80 & 0.89 & 0.8329 \\ 
$\mathcal{M}_P+$ & 0.84 & 0.82 & 0.59 & 0.97 & 0.97 & 0.95 & 0.88 & 0.68 & 0.84 & 0.93 & 0.81 & 0.89 & 0.8470 \\ 
\hline
\end{tabular}
\caption{Performance of our models under the $Acc_{\frac{\pi}{6}}$ metric (higher is better).}
\label{table:Acc30}
\end{table}

\begin{table}
	\centering
	\small
	\setlength{\tabcolsep}{1.2mm}
	\begin{tabular}{|c|cccccccccccc|c|}
		\hline
		Method & aero & bike & boat & bottle & bus & car & chair & dtable & mbike & sofa & train & tv & Mean \\
		\hline
		\multirow{2}{*}{K=24} & 12.4 & 16.3 & 23.5 & 8.7 & 2.7 & 5.4 & 11.6 & 17.4 & 16.3 & 13.7 & 6.2 & 15.6 & 12.48 \\ 
		& 0.84 & 0.82 & 0.58 & 0.95 & 0.97 & 0.94 & 0.89 & 0.57 & 0.77 & 0.91 & 0.81 & 0.86 & 0.8266 \\ 
		\hline
		\multirow{2}{*}{K=50} & 12.9 & 16.0 & 21.1 & 8.3 & 3.3 & 6.1 & 11.2 & 22.2 & 17.7 & 11.8 & 5.8 & 13.9 & 12.53 \\
		& 0.82 & 0.80 & 0.59 & 0.95 & 0.97 & 0.94 & 0.92 & 0.54 & 0.78 & 0.91 & 0.83 & 0.89 & 0.8281 \\  
		\hline
		\multirow{2}{*}{K=100} & 12.1 & 16.0 & 19.9 & 8.9 & 3.4 & 6.5 & 10.8 & 15.2 & 16.4 & 9.6 & 5.9 & 13.0 & 11.48 \\
		& 0.83 & 0.76 & 0.63 & 0.96 & 0.97 & 0.93 & 0.91 & 0.57 & 0.78 & 0.95 & 0.82 & 0.88 & 0.8335 \\ 
		\hline
		\multirow{2}{*}{K=200} & 10.6 & 16.4 & 21.6 & 8.1 & 3.2 & 6.0 & 9.9 & 14.6 & 16.0 & 11.1 & 6.3 & 13.4 & 11.44 \\ 
		& 0.84 & 0.76 & 0.62 & 0.96 & 0.98 & 0.94 & 0.92 & 0.65 & 0.80 & 0.96 & 0.82 & 0.87 & 0.8439 \\
		\hline
	\end{tabular}
	\caption{Ablation analysis of the size of K-Means dictionary in model $\mathcal{M}_G$ under the $MedErr$ (first row) and $Acc_{\frac{\pi}{6}}$ (second row) metrics.}
	\label{table:kmeans_m1}
\end{table}

\begin{table}
	\centering
	\small
	\setlength{\tabcolsep}{1.2mm}
	\begin{tabular}{|c|cccccccccccc|c|}
		\hline
		Method & aero & bike & boat & bottle & bus & car & chair & dtable & mbike & sofa & train & tv & Mean \\
		\hline
		\multirow{2}{*}{K=4} & 10.4 & 13.3 & 21.9 & 7.2 & 2.9 & 5.3 & 9.9 & 16.3 & 14.1 & 10.4 & 5.0 & 12.5 & 10.78 \\ 
		& 0.85 & 0.80 & 0.61 & 0.97 & 0.97 & 0.95 & 0.87 & 0.67 & 0.84 & 0.93 & 0.83 & 0.86 & 0.8453 \\ 
		\hline
		\multirow{2}{*}{K=8} & 10.5 & 14.8 & 21.5 & 6.8 & 2.7 & 4.9 & 9.7 & 16.1 & 14.9 & 10.2 & 5.6 & 12.3 & 10.85 \\
		& 0.83 & 0.79 & 0.60 & 0.97 & 0.96 & 0.95 & 0.91 & 0.62 & 0.81 & 0.95 & 0.83 & 0.89 & 0.8427 \\
		\hline
		\multirow{2}{*}{K=16} & 9.9 & 14.3 & 21.3 & 7.3 & 2.7 & 4.9 & 9.6 & 13.0 & 14.7 & 10.8 & 5.2 & 11.7 & 10.46 \\
		& 0.84 & 0.82 & 0.61 & 0.96 & 0.98 & 0.96 & 0.92 & 0.67 & 0.82 & 0.97 & 0.82 & 0.90 & 0.8553 \\
		\hline
		\multirow{2}{*}{K=24} & 9.7 & 15.3 & 23.5 & 7.1 & 2.9 & 5.0 & 10.0 & 13.3 & 14.4 & 11.3 & 5.3 & 13.1 & 10.91 \\
		& 0.87 & 0.80 & 0.60 & 0.96 & 0.97 & 0.95 & 0.90 & 0.65 & 0.83 & 0.94 & 0.82 & 0.87 & 0.8467 \\ 
		\hline
	\end{tabular}
	\caption{Ablation analysis of the size of K-Means dictionary in model $\mathcal{M}_G+$ under the $MedErr$ and $Acc_{\frac{\pi}{6}}$ metrics.}
	\label{table:kmeans_m1+}
\end{table}

\begin{table}
	\centering
	\small
	\setlength{\tabcolsep}{1.2mm}
	\begin{tabular}{|c|cccccccccccc|c|}
		\hline
		Method & aero & bike & boat & bottle & bus & car & chair & dtable & mbike & sofa & train & tv & Mean \\
		\hline
		\multirow{2}{*}{$\alpha=0.1$} & 11.8 & 16.1 & 20.8 & 8.4 & 3.3 & 6.4 & 10.6 & 28.5 & 15.0 & 11.2 & 6.0 & 12.3 & 12.53 \\
		& 0.84 & 0.77 & 0.62 & 0.96 & 0.96 & 0.94 & 0.91 & 0.51 & 0.82 & 0.96 & 0.81 & 0.88 & 0.8306 \\
		\hline
		\multirow{2}{*}{$\alpha=1$} & 12.1 & 16.0 & 19.9 & 8.9 & 3.4 & 6.5 & 10.8 & 15.2 & 16.4 & 9.6 & 5.9 & 13.0 & 11.48 \\
		& 0.83 & 0.76 & 0.63 & 0.96 & 0.97 & 0.93 & 0.91 & 0.57 & 0.78 & 0.95 & 0.82 & 0.88 & 0.8335 \\
		\hline
		\multirow{2}{*}{$\alpha=10$} & 12.1 & 14.5 & 22.8 & 8.7 & 3.1 & 6.5 & 10.9 & 15.1 & 16.3 & 10.6 & 6.0 & 13.0 & 11.63 \\ 
		& 0.82 & 0.79 & 0.59 & 0.96 & 0.97 & 0.94 & 0.91 & 0.67 & 0.81 & 0.95 & 0.82 & 0.88 & 0.8424 \\
		\hline
	\end{tabular}
	\caption{Ablation analysis of the weighting parameter $\alpha$ in model $\mathcal{M}_G$ under the $MedErr$ and $Acc_{\frac{\pi}{6}}$ metrics.}
	\label{table:alpha_m1}
\end{table}

\begin{table}
	\centering
	\small
	\setlength{\tabcolsep}{1.2mm}
	\begin{tabular}{|c|cccccccccccc|c|}
		\hline
		Method & aero & bike & boat & bottle & bus & car & chair & dtable & mbike & sofa & train & tv & Mean \\
		\hline
		\multirow{2}{*}{$\alpha=0.1$} & 10.3 & 16.0 & 24.0 & 7.1 & 3.2 & 5.5 & 10.3 & 11.8 & 15.2 & 10.6 & 6.0 & 12.3 & 11.01 \\
		& 0.86 & 0.81 & 0.59 & 0.96 & 0.98 & 0.95 & 0.90 & 0.65 & 0.80 & 0.96 & 0.82 & 0.89 & 0.8473 \\ 
		\hline
		\multirow{2}{*}{$\alpha=1$} & 9.9 & 14.3 & 21.3 & 7.3 & 2.7 & 4.9 & 9.6 & 13.0 & 14.7 & 10.8 & 5.2 & 11.7 & 10.46 \\
		& 0.84 & 0.82 & 0.61 & 0.96 & 0.98 & 0.96 & 0.92 & 0.67 & 0.82 & 0.97 & 0.82 & 0.90 & 0.8553 \\ 
		\hline
		\multirow{2}{*}{$\alpha=10$} & 8.5 & 14.8 & 20.5 & 7.0 & 3.1 & 5.1 & 9.3 & 11.3 & 14.2 & 10.2 & 5.6 & 11.7 & 10.10 \\
		& 0.87 & 0.81 & 0.64 & 0.96 & 0.97 & 0.95 & 0.92 & 0.67 & 0.85 & 0.97 & 0.82 & 0.88 & 0.8588 \\
		\hline
	\end{tabular}
	\caption{Ablation analysis of the weighting parameter $\alpha$ in model $\mathcal{M}_G+$ under the $MedErr$ and $Acc_{\frac{\pi}{6}}$ metrics.}
	\label{table:alpha_m1+}
\end{table}

\appendix

\section{Supplementary Results}
\label{sec:more_results}
\begin{figure}[h]
	\centering
	\subfigure{\includegraphics[width=\linewidth]{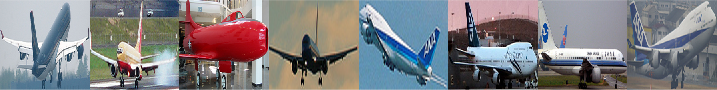}}
	\subfigure{\includegraphics[width=\linewidth]{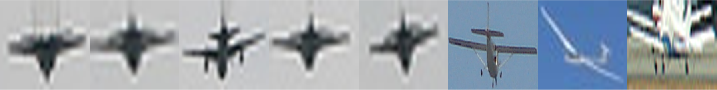}}
	\caption{Best (top row) and Worst (bottom row) images for Category: Aeroplane}
	\label{fig:aeroplane}
\end{figure}

\begin{figure}[h]
	\centering
	\subfigure{\includegraphics[width=\linewidth]{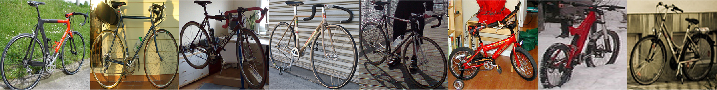}}
	\subfigure{\includegraphics[width=\linewidth]{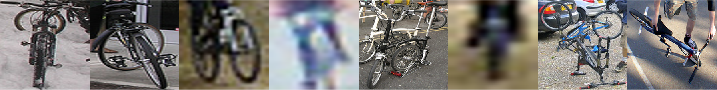}}
	\caption{Best (top row) and Worst (bottom row) images for Category: Bicycle}
	\label{fig:bicycle}
\end{figure}

\begin{figure}[h]
	\centering
	\subfigure{\includegraphics[width=\linewidth]{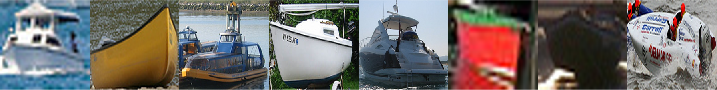}}
	\subfigure{\includegraphics[width=\linewidth]{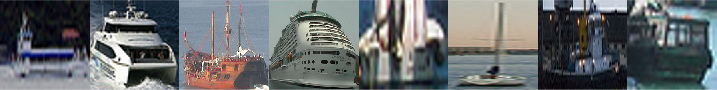}}
	\caption{Best (top row) and Worst (bottom row) images for Category: Boat}
	\label{fig:boat}
\end{figure}

\begin{figure}[h]
	\centering
	\subfigure{\includegraphics[width=\linewidth]{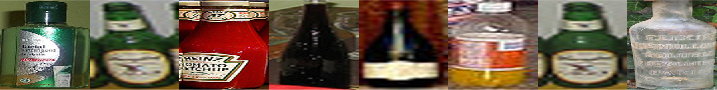}}
	\subfigure{\includegraphics[width=\linewidth]{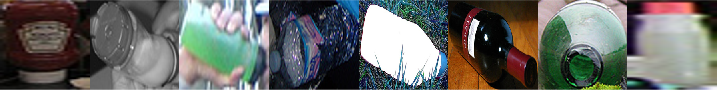}}
	\caption{Best (top row) and Worst (bottom row) images for Category: Bottle}
	\label{fig:bottle}
\end{figure}

\begin{figure}[h]
	\centering
	\subfigure{\includegraphics[width=\linewidth]{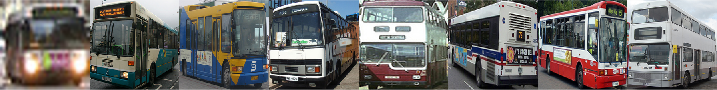}}
	\subfigure{\includegraphics[width=\linewidth]{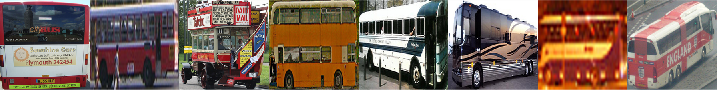}}
	\caption{Best (top row) and Worst (bottom row) images for Category: Bus}
	\label{fig:bus}
\end{figure}

\begin{figure}[h]
	\centering
	\subfigure{\includegraphics[width=\linewidth]{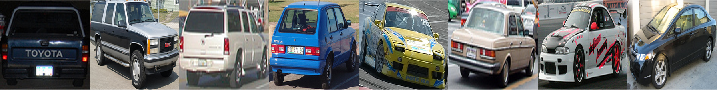}}
	\subfigure{\includegraphics[width=\linewidth]{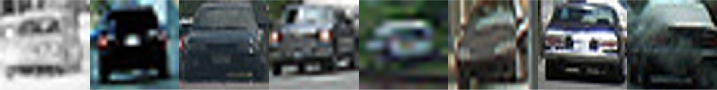}}
	\caption{Best (top row) and Worst (bottom row) images for Category: Car}
	\label{fig:car}
\end{figure}

\begin{figure}[h]
	\centering
	\subfigure{\includegraphics[width=\linewidth]{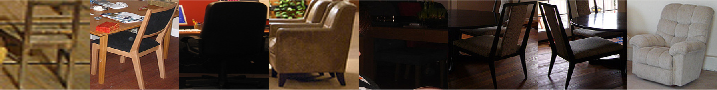}}
	\subfigure{\includegraphics[width=\linewidth]{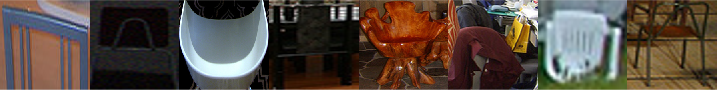}}
	\caption{Best (top row) and Worst (bottom row) images for Category: Chair}
	\label{fig:chair}
\end{figure}

\begin{figure}[h]
	\centering
	\subfigure{\includegraphics[width=\linewidth]{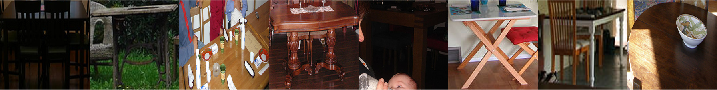}}
	\subfigure{\includegraphics[width=\linewidth]{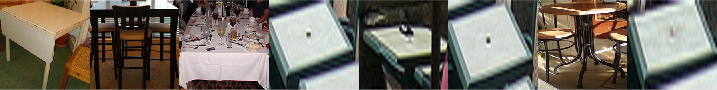}}
	\caption{Best (top row) and Worst (bottom row) images for Category: Diningtable}
	\label{fig:diningtable}
\end{figure}

\begin{figure}[h]
	\centering
	\subfigure{\includegraphics[width=\linewidth]{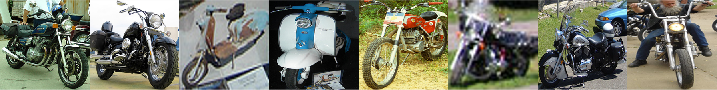}}
	\subfigure{\includegraphics[width=\linewidth]{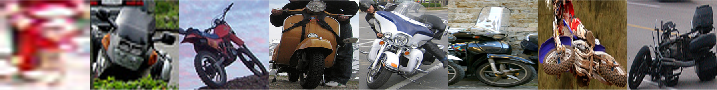}}
	\caption{Best (top row) and Worst (bottom row) images for Category: Motorbike}
	\label{fig:motorbike}
\end{figure}

\begin{figure}[h]
	\centering
	\subfigure{\includegraphics[width=\linewidth]{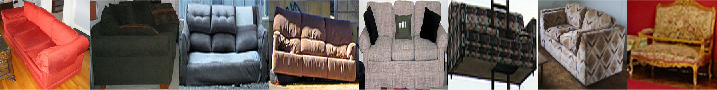}}
	\subfigure{\includegraphics[width=\linewidth]{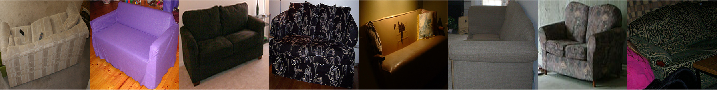}}
	\caption{Best (top row) and Worst (bottom row) images for Category: Sofa}
	\label{fig:sofa}
\end{figure}

\begin{figure}[h]
	\centering
	\subfigure{\includegraphics[width=\linewidth]{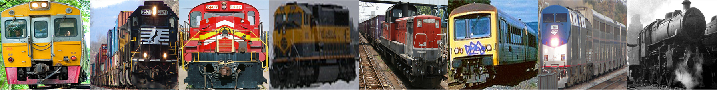}}
	\subfigure{\includegraphics[width=\linewidth]{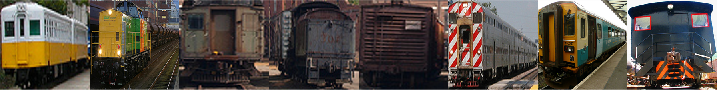}}
	\caption{Best (top row) and Worst (bottom row) images for Category: Train}
	\label{fig:train}
\end{figure}

\begin{figure}[h]
	\centering
	\subfigure{\includegraphics[width=\linewidth]{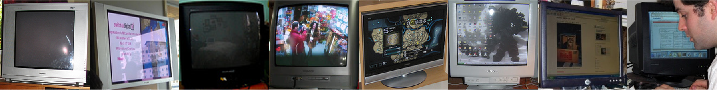}}
	\subfigure{\includegraphics[width=\linewidth]{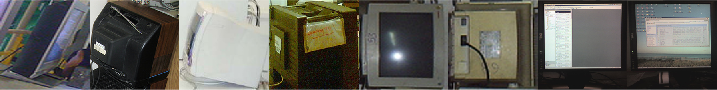}}
	\caption{Best (top row) and Worst (bottom row) images for Category: Tvmonitor}
	\label{fig:tvmonitor}
\end{figure}

In Figs.~\ref{fig:aeroplane}-\ref{fig:tvmonitor}, we show images of objects where we obtain the least pose estimation error and the most pose estimation error for every object category using model $\mathcal{M}_G+$. We make the most error under three conditions: (i) when the objects are really blurry (very small in pixel size in the original image), (ii) the shape of the object is uncommon (possibly very few examples seen during training) and (iii) the pose of a test image is very different from common poses observed during training. 
The first condition is best observed in the bad cases for categories aeroplane and car where almost all the images shown are very blurry. The second condition is best observed in categories boat and chair where the bad cases contain uncommon boats and chairs. The third condition is best observed in categories bottle and tvmonitor where the bad images are in very different poses compared to the best images. 

\section{Implementation details}
\label{sec:details}
For all our experiments, we train with a batch of 4 real and 4 rendered images across 12 object categories (which is a total of 96 images per batch). We train all our models using the Adam optimizer \cite{Kingma:ICLR15} with a starting learning rate of $10^{-4}$ and a reduction by a factor of 0.1 every epoch. The code was written using PyTorch \cite{Paszke:NIPS17}. We use the ResNet-50 upto layer4 (2048-dim feature output) as our feature network. The pose networks are of the form Input-FC-BN-ReLU-FC-BN-ReLU-FC-Output where FC is a fully connected layer, BN is a batch normalization layer and ReLU is the standard rectified linear unit non-linearity. The pose networks for models $\mathcal{R}_E$ and $\mathcal{R}_G$ are of size 2048-1000-500-3. The pose network of model $\mathcal{C}$ is of size 2048-1000-500-100 where 100 is the size of the K-Means dictionary we use to discretize the pose space. The bin and delta networks of models $\mathcal{M}_S$, $\mathcal{M}_G$, $\mathcal{M}_R$ and $\mathcal{M}_P$ are of sizes 2048-1000-500-100 and 2048-1000-500-3 respectively. For models $\mathcal{M}_S+$, $\mathcal{M}_G+$, $\mathcal{M}_R+$ and $\mathcal{M}_P+$ where we have one delta network per pose-bin per object category, our bin network is of size 2048-1000-500-16 (corresponding to 16 pose-bins) and we use a 2-layer delta network of size 2048-100-3. 

For the models, $\mathcal{M}_G$ and $\mathcal{M}_G+$, we initialize the network weights with 1 epoch of training over the models $\mathcal{M}_S$ and $\mathcal{M}_S+$. All other models are initialized  using pre-trained networks on the ImageNet image classification problem. The models $\mathcal{M}_S$, $\mathcal{M}_G$, $\mathcal{M}_P$, $\mathcal{M}_S+$ and $\mathcal{M}_P+$ were trained with $\alpha=1$. For the model $\mathcal{M}_G+$, we use a value of $\alpha=10$ and for the models $\mathcal{M}_R$ and $\mathcal{M}_R+$, we use $\alpha=0.1$.

\clearpage

\bibliography{recognition,vidal}

\begin{thebibliography}{25}
\providecommand{\natexlab}[1]{#1}
\providecommand{\url}[1]{\texttt{#1}}
\expandafter\ifx\csname urlstyle\endcsname\relax
  \providecommand{\doi}[1]{doi: #1}\else
  \providecommand{\doi}{doi: \begingroup \urlstyle{rm}\Url}\fi

\bibitem[PAS()]{PASCAL}
The {P}{A}{S}{C}{A}{L} {O}bject {R}ecognition {D}atabase {C}ollection.
\newblock \url{http://www.pascal-network.org/challenges/VOC/databases.html}.

\bibitem[Crivellaro et~al.(2015)Crivellaro, Rad, Verdie, Yi, Fua, and
  Lepetit]{Crivellaro:ICCV15}
Alberto Crivellaro, Mahdi Rad, Yannick Verdie, Kwang~Moo Yi, Pascal Fua, and
  Vincent Lepetit.
\newblock A novel representation of parts for accurate 3d object detection and
  tracking in monocular images.
\newblock In \emph{{IEEE} International Conference on Computer Vision}, 2015.

\bibitem[Deng et~al.(2009)Deng, Dong, Socher, Li, Li, and Fei-fei]{ImageNet}
Jia Deng, Wei Dong, Richard Socher, Li-Jia Li, Kai Li, and Li~Fei-fei.
\newblock Imagenet: A large-scale hierarchical image database.
\newblock In \emph{{IEEE} Conference on Computer Vision and Pattern
  Recognition}, 2009.

\bibitem[Elhoseiny et~al.(2016)Elhoseiny, El-Gaaly, Bakry, and
  Elgammal]{Elhoseiny:ICML16}
Mohamed Elhoseiny, Tarek El-Gaaly, Amr Bakry, and Ahmed Elgammal.
\newblock A comparative analysis and study of multiview {CNN} models for joint
  object categorization and pose estimation.
\newblock In \emph{International Conference on Machine learning}, 2016.

\bibitem[Grabner et~al.(2018)Grabner, Roth, and Lepetit]{Grabner:CVPR18}
Alexander Grabner, Peter~M. Roth, and Vincent Lepetit.
\newblock 3d pose estimation and 3d model retrieval for objects in the wild.
\newblock In \emph{{IEEE} Conference on Computer Vision and Pattern
  Recognition}, 2018.

\bibitem[G{\"{u}}ler et~al.(2017)G{\"{u}}ler, Trigeorgis, Antonakos, Snape,
  Zafeiriou, and Kokkinos]{Guler:CVPR17}
Riza~Alp G{\"{u}}ler, George Trigeorgis, Epameinondas Antonakos, Patrick Snape,
  Stefanos Zafeiriou, and Iasonas Kokkinos.
\newblock Densereg: Fully convolutional dense shape regression in-the-wild.
\newblock In \emph{{IEEE} Conference on Computer Vision and Pattern
  Recognition}, 2017.

\bibitem[G{\"{u}}ler et~al.(2018)G{\"{u}}ler, Trigeorgis, Antonakos, Snape,
  Zafeiriou, and Kokkinos]{Guler:arxiv18}
Riza~Alp G{\"{u}}ler, George Trigeorgis, Epameinondas Antonakos, Patrick Snape,
  Stefanos Zafeiriou, and Iasonas Kokkinos.
\newblock Densereg: Fully convolutional dense shape regression in-the-wild.
\newblock \emph{coRR abs/1803.02188}, 2018.
\newblock URL \url{http://arxiv.org/abs/1803.02188}.

\bibitem[He et~al.(2016{\natexlab{a}})He, Zhang, Ren, and Sun]{He:CVPR16}
Kaiming He, Xiangyu Zhang, Shaoqing Ren, and Jian Sun.
\newblock Deep residual learning for image recognition.
\newblock In \emph{{IEEE} Conference on Computer Vision and Pattern
  Recognition}, 2016{\natexlab{a}}.

\bibitem[He et~al.(2016{\natexlab{b}})He, Zhang, Ren, and Sun]{He:arxiv16}
Kaiming He, Xiangyu Zhang, Shaoqing Ren, and Jian Sun.
\newblock Identity mappings in deep residual networks.
\newblock \emph{CoRR abs/1603.05027}, 2016{\natexlab{b}}.
\newblock URL \url{http://arxiv.org/abs/1603.05027}.

\bibitem[Jordan and Jacobs(1994)]{Jordan:NC94}
Michael~I. Jordan and Robert~A. Jacobs.
\newblock Hierarchical mixtures of experts and the em algorithm.
\newblock \emph{Neural Computation}, 6\penalty0 (2):\penalty0 181--214, 1994.
\newblock \doi{10.1162/neco.1994.6.2.181}.
\newblock URL \url{https://doi.org/10.1162/neco.1994.6.2.181}.

\bibitem[Kingma and Ba(2015)]{Kingma:ICLR15}
Diederik~P. Kingma and Jimmy Ba.
\newblock Adam: A method for stochastic optimization.
\newblock In \emph{International Conference on Learning Representations}, 2015.

\bibitem[Li et~al.(2018)Li, Bai, and Hager]{Li:arxiv18}
Chi Li, Jin Bai, and Gregory~D. Hager.
\newblock A unified framework for multi-view multi-class object pose
  estimation.
\newblock \emph{coRR abs/1801.08103}, 2018.
\newblock URL \url{http://arxiv.org/abs/1801.08103}.

\bibitem[Mahendran et~al.(2017)Mahendran, Ali, and Vidal]{Mahendran:ICCVW17}
Siddharth Mahendran, Haider Ali, and Ren\'e Vidal.
\newblock {3D} pose regression using convolutional neural networks.
\newblock In \emph{{IEEE} International Conference on Computer Vision Workshop
  on Recovering {6D} Object Pose}, 2017.

\bibitem[Massa et~al.(2014)Massa, Aubry, and Marlet]{Massa:arxiv14}
Francisco Massa, Mathieu Aubry, and Renaud Marlet.
\newblock Convolutional neural networks for joint object detection and pose
  estimation: {A} comparative study.
\newblock \emph{CoRR abs/1412.7190}, 2014.
\newblock URL \url{http://arxiv.org/abs/1412.7190}.

\bibitem[Massa et~al.(2016)Massa, Marlet, and Aubry]{Massa:BMVC16}
Francisco Massa, Renaud Marlet, and Mathieu Aubry.
\newblock Crafting a multi-task {CNN} for viewpoint estimation.
\newblock In \emph{British Machine Vision Conference}, 2016.

\bibitem[Mousavian et~al.(2017)Mousavian, Anguelov, Flynn, and
  Kosecka]{Mousavian:CVPR17}
Arsalan Mousavian, Dragomir Anguelov, John Flynn, and Jana Kosecka.
\newblock 3d bounding box estimation using deep learning and geometry.
\newblock In \emph{CVPR}, 2017.

\bibitem[Paszke et~al.(2017)Paszke, Gross, Chintala, Chanan, Yang, DeVito, Lin,
  Desmaison, Antiga, and Lerer]{Paszke:NIPS17}
Adam Paszke, Sam Gross, Soumith Chintala, Gregory Chanan, Edward Yang, Zachary
  DeVito, Zeming Lin, Alban Desmaison, Luca Antiga, and Adam Lerer.
\newblock Automatic differentiation in pytorch.
\newblock In \emph{NIPS-W}, 2017.

\bibitem[Pavlakos et~al.(2017)Pavlakos, Zhou, Chan, Derpanis, and
  Daniilidis]{Pavlakos:ICRA17}
Georgios Pavlakos, Xiaowei Zhou, Aaron Chan, Konstantinos~G Derpanis, and
  Kostas Daniilidis.
\newblock {6-DoF} object pose from semantic keypoints.
\newblock In \emph{{IEEE} International Conference on Robotics and Automation},
  2017.

\bibitem[Rad and Lepetit(2017)]{Rad:ICCV17}
Mahdi Rad and Vincent Lepetit.
\newblock Bb8: A scalable, accurate, robust to partial occlusion method for
  predicting the 3d poses of challenging objects without using depth.
\newblock In \emph{{IEEE} International Conference on Computer Vision}, 2017.

\bibitem[Ren et~al.(2015)Ren, He, Girshick, and Sun]{Ren:FasterRCNN}
Shaoqing Ren, Kaiming He, Ross Girshick, and Jian Sun.
\newblock {Faster R-CNN}: Towards real-time object detection with region
  proposal networks.
\newblock \emph{arXiv preprint arXiv:1506.01497}, 2015.

\bibitem[Su et~al.(2015)Su, Qi, Li, and Guibas]{Su:ICCV15}
Hao Su, Charles~R. Qi, Yangyan Li, and Leonidas~J. Guibas.
\newblock Render for cnn: Viewpoint estimation in images using cnns trained
  with rendered 3d model views.
\newblock In \emph{{IEEE} International Conference on Computer Vision}, 2015.

\bibitem[Tulsiani and Malik(2015)]{Tulsiani:CVPR15}
Shubham Tulsiani and Jitendra Malik.
\newblock Viewpoints and keypoints.
\newblock In \emph{{IEEE} Conference on Computer Vision and Pattern
  Recognition}, 2015.

\bibitem[Wang et~al.(2016)Wang, Li, Jia, and Liang]{Wang:PCM16}
Yumeng Wang, Shuyang Li, Mengyao Jia, and Wei Liang.
\newblock Viewpoint estimation for objects with convolutional neural network
  trained on synthetic images.
\newblock In \emph{Advances in Multimedia Information Processing - PCM 2016},
  2016.

\bibitem[Wu et~al.(2016)Wu, Xue, Lim, Tian, Tenenbaum, Torralba, and
  Freeman]{Wu:ECCV16}
Jiajun Wu, Tianfan Xue, Joseph~J Lim, Yuandong Tian, Joshua~B Tenenbaum,
  Antonio Torralba, and William~T Freeman.
\newblock {Single Image 3D Interpreter Network}.
\newblock In \emph{European Conference on Computer Vision}, pages 365--382,
  2016.

\bibitem[Yu~Xiang and Savarese(2014)]{Xiang:WACV14}
Roozbeh~Mottaghi Yu~Xiang and Silvio Savarese.
\newblock Beyond {PASCAL}: A benchmark for {3D} object detection in the wild.
\newblock In \emph{IEEE Winter Conference on Applications of Computer Vision},
  2014.

\end{thebibliography}
\end{document}